\documentclass{article}

\usepackage[english]{babel}

\usepackage[letterpaper,top=2cm,bottom=2cm,left=3cm,right=3cm,marginparwidth=1.75cm]{geometry}

\usepackage{amsmath}
\usepackage{amssymb}
\usepackage{graphicx}
\usepackage[colorlinks=true, allcolors=blue]{hyperref}

\usepackage{subcaption}
\usepackage{cleveref}

\usepackage{authblk}

\title{Input Reconstruction Attack against Vertical Federated \\Large Language Models}
\author[ ]{Fei Zheng}
\affil[ ]{College of Computer Science and Technology, Zhejiang University}
\affil[ ]{\textit{zfscgy2@zju.edu.cn}}

\begin{document}
\maketitle

\begin{abstract}
Recently, \textit{large language models} (LLMs) have drawn extensive attention from academia and the public, due to the advent of the ChatGPT.
While LLMs show their astonishing ability in text generation for various tasks, privacy concerns limit their usage in real-life businesses.
More specifically, either the user's inputs (the user sends the query to the model-hosting server) or the model (the user downloads the complete model) itself will be revealed during the usage.
\textit{Vertical federated learning} (VFL) is a promising solution to this kind of problem.
It protects both the user's input and the knowledge of the model by splitting the model into a bottom part and a top part, which is maintained by the user and the model provider, respectively.
However, in this paper, we demonstrate that in LLMs, VFL fails to protect the user input since it is simple and cheap to reconstruct the input from the intermediate embeddings.
Experiments show that even with a commercial GPU, the input sentence can be reconstructed in only one second.
We also discuss several possible solutions to enhance the privacy of vertical federated LLMs.
\end{abstract}

\section{Introduction}
With the advent of ChatGPT~\cite{chatgpt}, \textit{large language models} (LLMs) have drawn much attention from both the public and academia.
LLMs have shown great ability in various tasks, e.g., question answering, reading comprehension, text summarization, and even mathematical reasoning.
However, in real applications, LLMs still face significant privacy concerns.
For example, a typical scenario of ChatGPT's usage is that, the user sends his query to the OpenAI server, then gets the response.
In this case, the user's query is totally exposed to the LLM provider.
This is unacceptable when the query contains sensitive or valuable information, e.g., the user's personal information or confidential data of the company.
Another scenario is that the user downloads the complete LLM model and deploys it himself.
Although user privacy is protected in this case, the model is completely revealed to the user, which violates the model provider's privacy considering LLMs are highly valuable assets.

The fundamental problem here is that the data and the model are held by two different parties and shall not be revealed to each other.
One way to tackle this problem is cryptographic methods.
These methods use cryptographic technologies such as multiparty computation protocols, garbled circuits, and homomorphic encryptions to implement secure neural network inference or training.
While these methods fully protect both parties' privacy in a strict sense, by now they can only be applied to relatively simple models, due to the expensive computation and communication overhead.
A few pioneering works apply cryptographic methods to realize secure transformers~\cite{haomeng2022iron_transformers,houxiaoyang2023ciphergpt,dongye2023puma}.
However, they require at least several minutes to generate one token with very short input and under ideal local network settings, using high-performance servers, which is impractical in real-life applications.

An alternative method to protect privacy in this case is \textit{vertical federated learning} (VFL).
The typical method for VFL is \textit{split learning}, whose procedure can be described as follows:
\begin{enumerate}
    \item the model provider splits the model by a certain layer to get a \textit{bottom model} and a \textit{top model}.
    He then sends the bottom part of the model (\textit{bottom model}) to the user.
    \item During inference, the user feeds his private input to the bottom model, whose output is the hidden embeddings (smashed data).
    Then he sends the hidden embeddings to the model provider, who feeds it to the top model and gets the model output.
    The model provider sends the model output to the user.
\end{enumerate}
Hence, in split learning, only the hidden embeddings and part of the model are revealed, which protects both parties' privacy to some degree.
Notice that, using split learning, the response to the user's query is also obtained by the model provider, which could also leak user's privacy.
However, in this paper, we focus on the privacy of user inputs rather than model outputs.

We show that the adversary can easily reconstruct the exact input text using the hidden embeddings and the parameters of the bottom model.
Specifically, we first reconstruct the word embeddings in the first layer, then we compute the similarity between the reconstructed word embeddings and the actual word embeddings in the bottom model, and find the most similar words.
This kind of attack can be very efficient, and the adversary does not need any side information except for the bottom model.
Such a setting is practical since we assume the adversary has compromised the model provider, who has the complete model.
The experiments are conducted on the open-sourced ChatGLM6B model.
We also investigated adding noise in the hidden embeddings before sending it to the model provider in order to enhance input privacy.
However, experiment results show that such defense is impractical, since the model performance is also severely downgraded.

\section{Related Work}
Although few studies investigated the input reconstruction in LLMs, reconstructing inputs from hidden embeddings is a long-studied problem in deep learning, and specifically, VFL.
It can be classified into two classes: black-box and white-box.
In the black-box scenario, the bottom model used to generate the embeddings is unknown, while in the white-box scenario, the adversary maintains the bottom model, including its structure and values of parameters.
In the black-box scenario, a basic reconstruction method is to train a reconstruction model that takes the hidden embeddings as input and outputs the original input~\cite{Vepakomma2020nopeek,hezecheng2019mia,songcongzheng2020embedding_leakage}.
Other methods include training a generator to invert~\cite{hezecheng2019mia,morris2023embedding_almost}, and training both surrogate models and inputs~\cite{erdogan2022unsplit}.
Although \cite{morris2023embedding_almost} also applies the inversion attack to language models, it adopts a black-box setting, i.e., the adversary has no information of the bottom model, while he can query it as many times as he wants.
This requires the adversary to train a generative model to reconstruct the inputs, which is time-consuming.
Instead, our paper focuses on the white-box setting, which is more practical.
As for the white-box scenario, the most straightforward method is to directly optimize the input, as the embedding is now differentiable~\cite{hezecheng2019mia,luoxinjian2021feature_inference_attack,songcongzheng2020embedding_leakage}.
However, the authors in \cite{songcongzheng2020embedding_leakage} directly optimize the one-hot word vector with Softmax relaxation, which leads to suboptimal performance since the dimension of the one-hot vector is too high.
Instead, we optimize the word embeddings first, then reconstruct the word according to the word embeddings.

\section{Preliminaries}
\subsection{Large Language Models}
\textit{Large language models} (LLMs) have gain substantial attention recently, due to their powerful ability in general-purpose language understanding and generation.
Nowadays, LLMs are mainly built upon the transformer architecture~\cite{vaswani2017attention,devlin2019bert} and trained with the autoregressive generation tasks.
They usually contain from billions to trillions parameters so that they are extremely expensive to train, and require high-performance GPUs for inference.
Notable examples of LLMs include OpenAI's GPT models (e.g., GPT~\cite{radford2018gpt}, GPT-2~\cite{radford2019gpt2}, GPT-3~\cite{brown2020gpt3}, GPT-3.5 and GPT-4, used in ChatGPT~\cite{chatgpt}), Google's PaLM (used in Bard), Meta's LLaMa~\cite{touvron2023llama}, and ChatGLM~\cite{duzhengxiao2022glm}.

\subsection{Vertical Federated Learning}
\textit{Vertical federated learning} (VFL) refers to the federated learning case such that the data is vertically distributed, i.e., different parties have different features columns (including label) of the same samples.
A typical technique for VFL is \textit{split learning}~\cite{vepakomma2018split_health}, in which multiple parties split the complete model into multiple parts.
During the inference, parties exchange intermediate results in order to get the model output.
The simplest split learning case is that one party ($P_0$) has the bottom model ($M_b$) and the other has the top part of the model ($M_t$).
During inference, $P_0$ computes the hidden embedding $h = M_b(x)$ and sends it to $P_1$.
$P_1$ computes $y = M_t(h)$ to get the model prediction.
In our case, the model provider sends the bottom model to the user, which consists of embedding layers and several transformer layers, while the top model is made of the rest of the transformer layers and the linear layer used to predict the next token.
This setting is realistic, since the model provider has the complete model, and it is usually not possible for the user and the model provider to collaboratively train a split LLM.

\section{Input Reconstruction from Embeddings}
The inference procedure of a large language model can be expressed as follows:
\begin{equation}
    \text{LLM}(x) = \text{Linear}(\text{MultiheadAttns}[\text{WordEmbedding}(x) + \text{PosEmbedding}(x)]),
\end{equation}
i.e., the input (tokenization) first goes through the word embedding layer and the positional embedding layer to get the first hidden embedding, and then it goes through multiple multi-head attention layers to get the final hidden state, after which a linear layer is used to predict logits.
In split learning, the bottom model contains the initial embedding layers and several multi-head attention layers, and the top model contains the rest of the multi-head attention layers and the final linear layer:
\begin{equation}
    \begin{cases}
        h_s = \text{LLM}_\text{bottom}(x) = \text{MultiheadAttns}_\text{bottom}[\text{WordEmbedding}(x) + \text{PosEmbedding}(x)] 
        \\
        y = \text{LLM}_\text{top}(h_s) = \text{Linear}[\text{MultiheadAttns}_\text{top}(h_s)]
    \end{cases}
\end{equation}

The input reconstruction attack can then be expressed as given $\text{LLM}_\text{bottom}(x)$, find a $x'$ such that $\text{LLM}_\text{bottom}(x') = \text{LLM}_\text{bottom}(x)$.
We divide this task into two steps: word embedding reconstruction and token reconstruction.

\subsection{Word Embedding Reconstruction}
The first step is to reconstruct the word embedding $h_w = \text{WordEmbedding}(x)$.
Given that the hidden embeddings have the same sequence length as the input, we can determine the positional embeddings (denoted as $h_p$) directly from the hidden embeddings:
\begin{equation}
\label{eq:embedding-reconstruct}
    \text{argmin}_{h'_w} \Vert \text{MultiheadAttns}_\text{bottom}(h'_w + h_p) - \text{MultiheadAttns}_\text{bottom}(h_w + h_p) \Vert^2.
\end{equation}
For large language models, all the hidden embeddings have the same size, which indicates that in \eqref{eq:embedding-reconstruct}, the number of variables is the same as the number of equations.
This is preferable to us since that $h'_w$ is less likely to have infinite solutions than in the case that the number of variables is less than the number of equations.
We use the gradient descent method to solve \eqref{eq:embedding-reconstruct}.
The initialization of $h_w'$ is also important, as we empirically found that a random initialization can cause computation errors when run in half precision.
Hence, we use actual sentences to generate the embedding initialization.
For example, if the sequence length is $n$, we can use $\text{WordEmbedding}(\text{`hello'}\times n)$ as the initial embedding.

\subsection{Token Reconstruction}
After the word embeddings $h'_w = (w'_1, \cdots, w'_l)$ is reconstructed, for each $w'_i$ we find the token with the most similar embedding in the word embedding table, i.e.,
\begin{equation}
    t'_i = \text{argmax}_{t'} \dfrac{\text{WordEmbedding}(t') \cdot w'_i}{\Vert\text{WordEmbedding}(t')\Vert \cdot \Vert w'_i \Vert}.
\end{equation}
Thus, the reconstructed input text is $(t'_1, \cdots, t'_s)$.

\subsection{Why not Reconstruct Tokens Directly?}
In \cite{songcongzheng2020embedding_leakage}, the authors directly reconstruct tokens from embedding by a token selection vector $v = (v_1, \cdots, v_D)$, where $D$ is the vocabulary size.
The word embedding then becomes a weighted sum: $h \in \mathbb R^d = \dfrac{1}{\sum_{j=1}^D e^{v_j}} \sum_{i} w_ie^{v_i}$, where Softmax is to ensure the sum of weights is 1, and $d$ is the dimension of word embedding.
However, notice that $D$ could be as large as 100K, and in most models, $d$ is much smaller than $D$.
This leads infinite solutions of the equation $\dfrac{1}{{\sum_{j=1}^D} e^{v_j}} \sum_{i} w_ie^{v_i} = h$, contradictory to the fact that $v$ is one-hot.

\subsection{Adding Noise for Defense}
A straightforward and commonly used defense strategy is adding noise to the hidden embeddings before sending it to the model provider.
The most common type of noise is Gaussian noise, which is widely used in deep learning to achieve differential privacy~\cite{abadi2016dpsgd,geyer2017dpfl,morris2023embedding_almost}.
Hence, we also investigate adding Gaussian noise to the hidden embeddings to protect privacy, i.e., for each element $h_i$ in hidden embeddings $h$, we add a random Gaussian variable to it: $\tilde h_i = h_i + \epsilon, \epsilon \sim \mathcal N(0, \sigma^2)$.
The standard deviation term $\sigma$ is used to control the scale of the Gaussian noise.

\section{Experiments}
We choose the ChatGLM6B model~\cite{zeng2022glm130b} for our experiments, and all experiments are run on the server with Nividia RTX 3090 GPUs.
We use the SQuAD QA (Question-Answering) dataset~\cite{2016squad} to evaluate the reconstruction accuracy and the model performance.
For the QA task, we use the greedy generation.
The input is the concatenated context and question, with an average number of tokens around 220.
For the reconstruction attack, we perform gradient descent until 1000 steps or the cosine similarity between the reconstructed hidden embedding reaches 0.98.
To study the attack performance under different input lengths, we also use the FindSum dataset~\cite{2022findsum}.
Here we only reconstruct the input based on the first query's hidden embeddings.

We use accuracy as the metric for both the QA task and the reconstruction attack.
In the QA task, we view a response as correct if the ground-truth answer string is contained in the models' response.
In the reconstruction attack, the accuracy is computed in a token-wise manner:
\begin{equation}
    \text{attack accuracy} = \dfrac{1}{N} \sum_{i=1}^N \dfrac{1}{l_i}\sum_{j=1}^{l_i} 1_{t_{i, j}=t'_{i,j}},
\end{equation}
where $t_{i, 1}, \cdots, t_{i, l_i}$ is the $i$-th input sequence, and $t'_{i, 1}, \cdots, t'_{i, l_i}$ is the sequence reconstructed by the attacker, and $1_{t=t'}$ is the indicator function.

\subsection{Different Text Lengths}
\begin{figure}[ht]
    \centering
    \begin{subfigure}{0.46\textwidth}
        \includegraphics[width=1\textwidth]{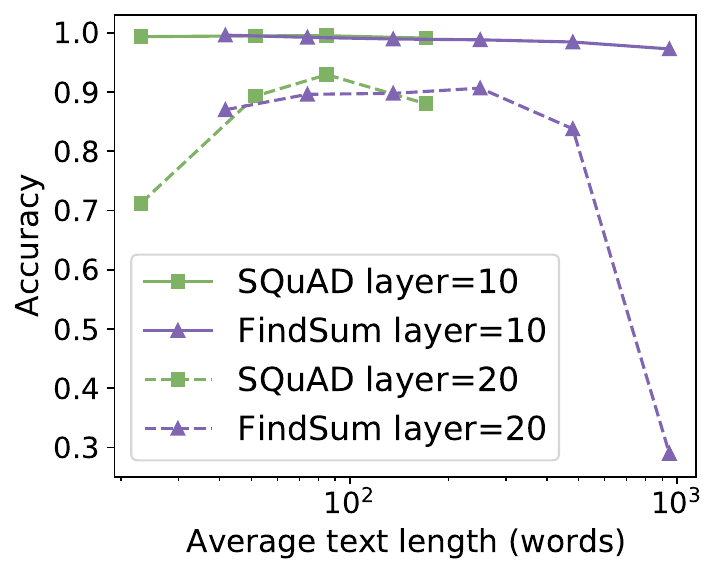}
        \caption{Attack accuracy vs. input length.}
    \end{subfigure}
    \begin{subfigure}{0.46\textwidth}
        \includegraphics[width=1\textwidth]{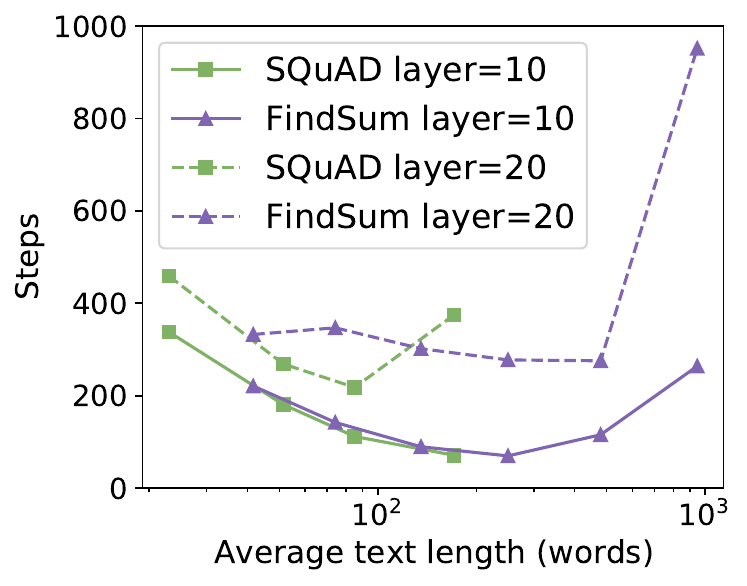}
        \caption{Recontruction steps vs. input length.}
    \end{subfigure}
    \caption{Reconstruction attack results with different input length.}
    \label{fig:seqlen}
\end{figure}

We report the attack accuracy and the number of gradient descent steps taken for different input lengths in \Cref{fig:seqlen}.
When splitting at the 10th layer, the attack accuracy is constantly around 100\% regardless of the input length.
On the other side, when splitting at the 20th layer, higher attack accuracy appears on middle-sized inputs (length between 50 and 500).
The attack accuracy drops to around 30\% when the input length is around 1,000.
The number of steps for the reconstruction attack shows a similar trend with the accuracy, i.e., fewer steps are needed for middle-sized inputs, and splitting at the 20th layer requires significantly more steps than at the 10th layer.

Notably, in our experiments, the maximum number of steps is set to 1,000. Thus the attack accuracy could increase with more reconstruction steps.

\subsection{Different Noise Levels}
\begin{figure}[ht]
    \centering
    \begin{subfigure}{0.46\textwidth}
        \includegraphics[width=1\textwidth]{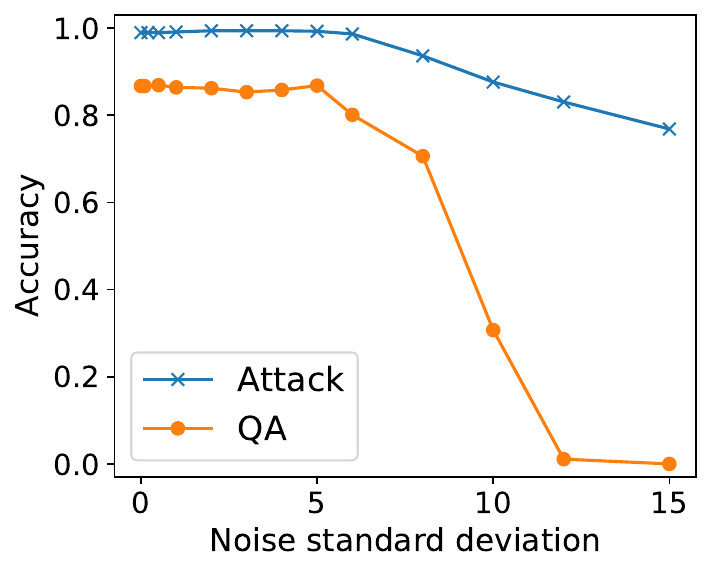}
        \caption{Split layer=10}
    \end{subfigure}
    \begin{subfigure}{0.46\textwidth}
        \includegraphics[width=1\textwidth]{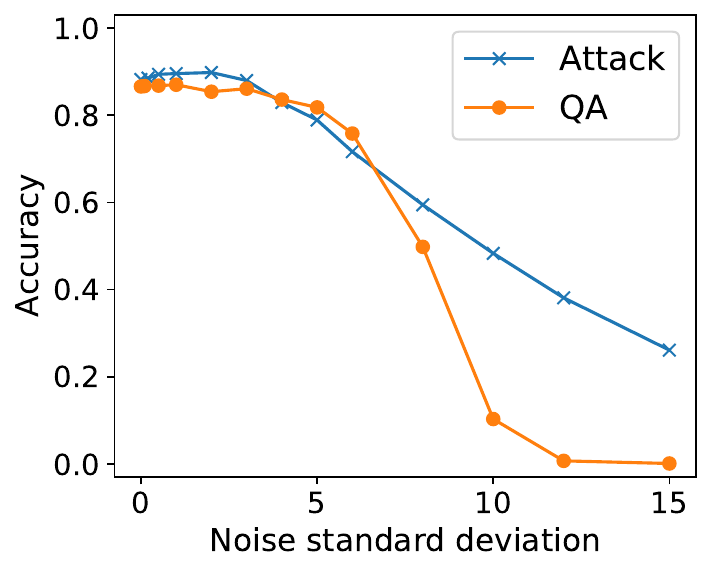}
        \caption{Split layer=20}
    \end{subfigure}
    \caption{QA performance vs. reconstruction performance, with different $\sigma$ (noise std.).}
    \label{fig:qa-rec}
\end{figure}

We show the QA accuracy and the attack accuracy under different noise levels in \Cref{fig:qa-rec}.
The performance of the QA task drops significantly when $\sigma$ reaches 5, and becomes zero at $\sigma=15$.
On the contrary, the accuracy of the reconstruction attack is close to 100\% at first and gradually decreases as $\sigma$ increases.
When we split the model at higher layers, we did observe a drop in the attack accuracy.
However, even splitting at the 20th layer (which already reveals about 2/3 of total model parameters), the attack accuracy can be still around 50\% when the QA accuracy becomes around zero.
Hence, we can conclude that simply adding noise in the forward embeddings is not a valid measure to protect input privacy.
The large portion of input can still be recovered even when the noise already totally damages the model utility.
This is unlike the black-box scenario in \cite{morris2023embedding_almost}, where the attack performance drops faster than the original task performance.

\section{Possible Defenses}
\subsection{New Backbone Architecture}
One reason for the easy reconstruction of input is that, the hidden embeddings are too large and contain too much information in the transformer backbone of LLMs.
For example, hidden embeddings in different layers have the same size as in the first layer, which enables the attacker to solve the equations.
Hence, the defense of input reconstruction attack may require a re-design of the current LLM backbone.
For example, shrinking the size of hidden embeddings layer by layer, or gradually dropping embeddings of early tokens.
However, these methods require a modification in the LLM backbone, and the verification can be very expensive.

\subsection{Split at Training}
The reconstruction attack is effective partly because of the white-box setting.
Hence, it is also possible to train the LLM in the split learning manner at the beginning, so that no party can obtain the complete model.
However, such methods require that the parties involved in split training do not collude with each other. 
Also, potential black-box attacks such like~\cite{morris2023embedding_almost} are possible against such defenses.-

\section{Conclusion}
In this paper, we investigate the possibility of applying vertical federated learning to large language models in a white-box scenario, i.e., the model is trained by the model provider but deployed using split learning.
We demonstrate that it is rather simple and costless for the attacker to reconstruct text from the hidden embeddings in a white-box scenario. 
The conventional noise-adding method fails to protect input privacy due to the model utility will be totally damaged before the noise defense starts to work.
Hence, to realize vertical federated large language models is still a challenging problem.
We also provide several possible solutions to enhance privacy for vertical federated LLMs.

\bibliographystyle{plain}
\bibliography{refs}

\end{document}